\documentclass[twoside,11pt]{article}
\usepackage[preprint]{jmlr2e}
\usepackage{hyperref}
\usepackage{booktabs}
\usepackage{verbatim}
\usepackage{longtable}
\usepackage{array}

\title{No Dataset Needed for Downstream Knowledge Benchmarking:
Response Dispersion Inversely Correlates with Accuracy on Domain-specific QA}
\author{Robert L Simione II\thanks{robert.simione@columbia.edu} \\ \texttt{Columbia University}}
\date{\today}
\editor{}

\begin{document}

\maketitle

\begin{abstract}
This research seeks to obviate the need for creating QA datasets and grading (chatbot) LLM responses when comparing LLMs' knowledge in specific topic domains. This is done in an entirely end-user centric way without need for access to any inner workings of the LLM, so long as it can be prompted and given a random seed to create different generations to the same prompt. The paper does this by, for a given topic domain, defining the "response dispersion" of an LLM by repeatedly asking an LLM the same opinion question about that topic domain. Namely, the response dispersion is the count of singular values needed to explain 95\% of the variance in the embedding matrix of the LLM's responses. It is found that the response dispersion is inversely correlated with accuracy on relevant QA evaluations (average spearman rank correlation stronger than -.59). A use-case analysis shows that when comparing two different LLMs on the same topic domain, comparing their response dispersion is a suitable replacement for comparing their QA accuracy between 74\% and 89\% of the time, the range depending on certain reasonable accuracy-difference tolerances that may be acceptable to an end-user in exchange for the labor being saved using response dispersion instead of QA accuracy for comparison. Two response embeddings are studied for creating the embedding matrix in this study, one is from OpenAI's APIs and one is a novel embedding, here named reference sentence similarity embeddings, that can be computed locally and performs very nearly as well in calculating response dispersion. Also in this research, a pre-existing dataset called the IRC-Wiki Trivia dataset, originally developed for trivia games, has been re-purposed, curated, and the curation, called \texttt{IRC-WikiTriviaQA}, is made available for the purpose of this research.  
\end{abstract}

\section{Introduction}

\subsection{Background and Motivation}\label{background-and-motivation}

When designing an AI-powered application that will use an LLM as part of its processing, a decision needs to be made about which LLM to use for the purpose. There are many benchmarks for comparing the capabilities of LLMs, and, other considerations held equal, it is generally preferable to use an LLM that has more knowledge about relevant topic domains than not. \citep[Section 4.2]{chang_survey_2024} While this seems true just intuitively, there is also experimental evidence verifying this. \citep[Section 2.3]{liu_lost_2024} showed that a model's ability to answer questions without any assistance can, in some circumstances, provide a correct answer more often than even when the answer to the question is itself a part of the prompt.\footnote{"For example, GPT-3.5-Turbo’s multi-document QA performance can drop by more than 20\%—in the worst case, performance in 20- and 30-document settings is lower than performance without any input documents (i.e., closed-book performance; 56.1\%)." \citep{liu_lost_2024}} As stated in \citep[Section 4.2]{chang_survey_2024}, "Question-answering benchmarks have become a fundamental component in the assessment of LLMs and their overall performance [on downstream tasks]."

However, creating datasets for QA benchmarking in a particular topic domain is labor intensive since an expert first needs to create a QA dataset relevant to the topic domain and then needs to grade LLM responses against the answer key. While the percentage of questions answered correctly is useful as a measurement of an LLM's knowledge in that topic domain, the goal of the practitioner building an AI-powered application under a deadline is merely to figure out which is the best currently existing LLM to plug into his or her application. This paper introduces a methodology where the practitioner can answer this question without creating a QA benchmarking dataset yet arrive at the same conclusion quickly and cheaply. In the scenario of comparing two LLM's against each other in a particular topic domain, 74\% of the time this novel methodology produces the same answer as would have been produced by creating the QA benchmark dataset and grading the LLM responses. Further, in exchange for the massive labor savings this process would give the practitioner, it is likely they would be willing to select the worse of the pair so long as it's not worse by "too much", quantified here by what will be called a "tolerance level". At a 5\% tolerance this procedure produces the better or good enough LLM 82\% of the time, and at 10\% tolerance the LLM selected is better or good enough 89\% of the time. 

\subsection{End-User Centric Assumptions}\label{end-user-centric}

The procedure created is designed with certain assumptions about a working practitioner in mind. This is because while some LLMs such as Meta's LLama series of models are open-source and provide access to their weights so that activations can be analyzed directly, it's also true that many LLMs used in practice are proprietary and thus kept behind remote-access-only API endpoints where neuron-activation-analysis is not possible. Thus in designing this procedure only 3 assumptions are made:
\begin{enumerate}
\item The chatbot is known to use a probabilistic procedure such that responses can vary depending on a randomizer seed value. (This is true of arguably all LLMs being used today but wouldn't be true of an older chatbot like ELIZA which may have zero response dispersion but would always perform poorly on a QA benchmark dataset.)
\item This random seed can be set so that multiple distinct generations to the same prompt can be accumulated.
\item The chatbot can be asked a question from a clean context window multiple times.
\end{enumerate}

All of these assumptions are valid for nearly all chatbots on the open market today. Specifically, this study tested LLMs available through OpenRouter.ai APIs, a platform on which many proprietary and open-source LLMs are available for prompting through an API endpoint. 

These end-user centric assumptions ensure that the procedure is applicable to a wide range of LLMs, including those that are proprietary and accessible only through API endpoints, thereby broadening the procedure's utility across different use cases.

\subsection{Procedure Overview}

The procedure detailed and validated in this paper is as follows:
\begin{enumerate}
    \item Specify your topic domain, e.g. "Sports", "Movies", "Music"
    \item For each LLM, ask it multiple times (in this paper 100) with different seeds to answer an opinion question about that topic domain to create a list of responses from that LLM.
    \item For each list of responses, create an embedding matrix where every row is the embedding of a response from the list. (Two different embedding methods are tested in this paper.)
    \item For each LLM's embedding matrix, count how many singular values are needed to explain 95\% of the variance in the matrix's rows. This count is here named the LLM's response dispersion for the specified topic domain.
    \item The LLM with the lower response dispersion either would perform better than the other LLM on a hypothetical QA benchmark dataset 74\% of the time, or it would perform within at-most 10\% tolerance on the hypothetical QA benchmark of the other LLM 89\% of the time.
\end{enumerate}

\subsection{Paper structure}

The rest of this paper is structured as follows: 
\begin{description}
\item[Section \ref{response-dispersion}] motivates the study of response dispersion and its relation to QA benchmark accuracy. Subsection \ref{motivating-hypothesis} describes the motivating hypothesis for why this measurement was focused on and studied with the expectation that it would correlate with QA benchmark accuracy. Subsection \ref{defining-response-dispersion-using-response-embeddings} defines response dispersion given an embedding matrix. Subsection \ref{response-embeddings-used} discusses the two text embeddings used and introduces a novel text embedding, reference sentence similarity embeddings, which on this task performed nearly as well as OpenAI's \texttt{text-embedding-3-large} embeddings while being faster and cheaper and able to compute locally.
\item[Section \ref{validation-methodology}] describes how the use of response dispersion was validated as useful for the intended LLM-comaring use-case. Subsection \ref{introducing-the-irc-wiki-trivia-dataset} introduces a repurposed and curated QA dataset of trivia questions across different topic domains called \texttt{IRC-WikiTriviaQA}. Subsection \ref{prompting-llm-responses-to-the-irc-wiki-trivia-questions} and \autoref{grading-the-llm-responses-to-the-irc-wiki-trivia-questions} describe how the \texttt{IRC-WikiTriviaQA} was used to find QA benchmark measurements for candidate LLMs studied for 12 different topic domains with \autoref{prompting-llm-responses-to-the-irc-wiki-trivia-questions} describing how LLMs were prompted to respond to \texttt{IRC-WikiTriviaQA} and \autoref{grading-the-llm-responses-to-the-irc-wiki-trivia-questions} describing and validating how the LLMs' responses were graded against the \texttt{IRC-WikiTriviaQA} answer key.

\item[Section \ref{results}] compiles and compares the QA benchmark scores collected with the response dispersions across the various topic domains in the \texttt{IRC-WikiTriviaQA} dataset. The results of the use-case analysis are presented, first averaged over all categories, then a table is presented for each category ordering the models studied on their QA benchmark accuracy, and on each of the response dispersions studied. 
\item[Section \ref{discussion-and-future}] discusses some shortcomings and not-yet-explored directions of the current work.
\end{description}

\section{Response Dispersion}\label{response-dispersion}

\subsection{Motivating Hypothesis}\label{motivating-hypothesis}

This section motivates the hypothesis that is later validated
experimentally.

Probabilistic text generation is a stochastic process guided by the
weights learned during training, where subsequent tokens in a response
are determined by logits output by an LLM. These logits are used to
define a probability distribution from which the next token of a
LLM's reply is selected. This means that different response
generations to the same prompt will generate different responses.

The motivating hypothesis of this paper is that sharper contrast in the
output logits of the LLM should correlate with the LLM being more sure
(measured by probability) in its output, which ought to (assuming the
information it was trained on is truthful) correlate to it having more
factual knowledge. However, since the goal is to measure this from the
point of view of an end-user of an LLM, it is not always possible to
measure the logits directly, as in the case with proprietary LLMs that
are accessed remotely through API endpoints. Thus this paper analyzes
ways to measure how dispersed the responses of an LLM are
when given the same prompt about a topic many times.

In an effort to give the latent diversity of potential responses the
best opportunity to be detected, a prompt template is used that asks the
LLM a generic opinion question about a topic category. The topic
category is left as a variable so that different domains of discourse
can be evaluated by plugging in their name into the \texttt{category}
variable. In particular, the prompt is:

\begin{quote}
Here is a test for evaluating LLMs. I want to see how well you follow my instructions when constructing your response.

I want you to respond with a SINGLE WORD, and ONLY THAT. Do not add any other context, other words, notes, explanations, justifications, or objections to my phrasing of this question. Now, please tell me in a single word what is your favorite thing to discuss related to the topic category of "\{category\}". Do not response with the "favorite", "discussions", or "\{category\}".
\end{quote}

This was asked of each LLM studied 100 times, with different passed as
an argument seeds for each ask. The explicit request to keep the answer
succint is there to help in measuring the diversity of answers by
keeping most of the response, because it is easier to programmatically
measure the difference in answers like {[}``XYZ'', ``XYZ'', ``ABC''{]}
rather than in answers like {[}``I like to talk about XYZ'', ``XYZ is
nice to discuss'', ``I like to talk about ABC''{]}.

    \subsection{Defining Response Dispersion Using Response
Embeddings}\label{defining-response-dispersion-using-response-embeddings}

The previous subsection motivated the need to measure how dispersed
responses to the prompt are when asked multiple times. Assuming you have
embeddings for each response that sufficiently capture the important
similarities and differences in responses to the prompt, then a
straightforward way to define response dispersion is to take the
embedding matrix where each row is the embedding of a response, and
count how many singular values of this matrix (starting from the largest
and counting in descending order) are required to explain some threshold
of variation in the embeddings, say 95\%.

    \subsection{Response Embeddings Used}\label{response-embeddings-used}

Different embedding methods can be dropped-in-place for defining the
Response Dispersion as above. This paper looks at two different
candidate embedding methods for embedding LLM responses. One is
simply to use the embeddings provided by OpenAI's text embedding API
using the model \texttt{text-embedding-3-large}, currently available
over a remotely accessed API endpoint. The other embedding method is
novel to this paper, here called ``reference sentence similarity
embeddings'' and abbreviated as rss embeddings. In the use-case analysis
described later, their performance is nearly indistinguishable with the
\texttt{text-embedding-3-large} embeddings having a marginally better
performance compared to rss embeddings; however rss embeddings have
benefits in that they capture almost all the same variation while being
lower dimension, they are quicker to obtain (compared to retrieving
embeddings over a remote API), and are computed locally.

Given a set of sentences to be called ``reference sentences''
\(\{r_{j}\}_{j}\), and a sentence similarity scoring method
\(s(\cdot, \cdot) \rightarrow [0,1]\), then a reference sentence
similarity embedding for a string \(t\) is a vector where the \(j\)-th
component of the embedding of \(t\) is \(s(t, r_{j})\).

From now on, this paper will use a normalized edit similarity score
called the normalized Indel similarity which is a special case of a
normalized Levenshtein similarity score.

The reference sentences used will be all the responses to be embedded.
This means that for a collection of responses \(\{t_{i}\}_i\) the
embedding matrix is going to its \(i,j\)-th component be
\(s(t_{i}, t_{j})\). Thus the \(i\)-th row of the embedding matrix will
be the rss embedding of \(t_i\) using all \(\{t_{j}\}_{j}\) as reference
sentences.

\section{Validation Methodology}\label{validation-methodology}

\subsection{Introducing the \texttt{IRC-WikiTriviaQA} Dataset}\label{introducing-the-irc-wiki-trivia-dataset}

To validate the use of response dispersion in comparing LLM's knowledge,
this paper presents a use-case analysis where the comparison of response
dispersion measurements is evaluated in its suitability as a proxy for
comparing between candidate LLMs' percentage of correct answers to
trivia questions, calculated on a per-category basis i.e. the result of
the gold-standard human labeling assessment described in the
introduction. The hand-labeled dataset used, \texttt{IRC-WikiTriviaQA}, is a repurposement and
curation of the original IRC-Wiki Trivia dataset which is available at \citep{bertrum_categorytrivia_2012}
and was released under a CC-by-SA-3.0 license \citep{creative_commons_deed_nodate}. The
author's curation of the dataset used for this paper is being released under the same license at \citep{simione_ii_no_2024}. The author kindly requests any users of \texttt{IRC-WikiTriviaQA} to cite both this paper and the original IRC-Wiki Trivia dataset.

The original dataset contains the categories "Animals", "Computers", "Food", "Geography", "History", "Movies", "Movies - Quote", "Music - Name the artist", "Music - Name the movie", "Music - Finish these lyrics", "Music", "Science", "Football", "Sport", "Religion/Mythology", "Mythology", "TV-Cartoons", and "TV" and 1671 questions. To curate the dataset, the following steps were taken:
\begin{enumerate}
    \item Any question from the "Anime", "Videogame", "Religion/Mythology", and "Mythology" categories was dropped. 
    \item The category "Movies - Quote" was merged into "Movies". 
    \item The categories "Music - Name the movie", "Music - Name the artist", and "Music - Finish these lyrics" were merged into "Music".
    \item Any question with multiple possible answers (as defined by the answer key) were dropped.
    \item Everything else is otherwise left the same.
\end{enumerate}
After this, the curated dataset has 1397 questions over 11 categories, broken down in the following table:
\begin{center}
\begin{tabular}{|c|r|}
\hline
\multicolumn{2}{|c|}{\texttt{IRC-WikiTriviaQA} breakdown} \\ \hline
\textbf{Category} & \textbf{\# of Questions} \\ \hline
Animals & 136 \\ \hline
Computers & 29 \\ \hline
Food & 218 \\ \hline
Football & 169 \\ \hline
Geography & 77 \\ \hline
History & 84 \\ \hline
Movies & 219 \\ \hline
Music & 287 \\ \hline
Science & 55 \\ \hline
Sport & 14 \\ \hline
TV & 17 \\ \hline
TV-Cartoons & 69 \\ \hline
\end{tabular}
\end{center}

\subsection{Prompting LLM responses to the IRC-Wiki Trivia
questions}\label{prompting-llm-responses-to-the-irc-wiki-trivia-questions}

This study evaluated 10 different chat-tuned foundation models against this
trivia dataset. For each trivia question in the dataset, the question
and its category (given by the dataset) were variables placed into the
following prompt template that was sent to each LLM studied:

\begin{quote}
Here is a test for evaluating LLMs. I want to see how well you follow my instructions when constructing your response.

I want you to respond with the answer to a "\{category\}" related trivia question, and ONLY THAT. Do not add any other context or explanation. Now, please tell me the answer to the following trivia question:"\{question\}"
\end{quote}

For each question, each LLM was prompted for a response once using a
random seed of 0 with a temperature of 0.\footnote{Note the use of a specific
temperature does not contradict the end-user-centric assumptions in subsection \ref{end-user-centric} because the end-user is not meant to be replicating this ground-truth
accuracy measurement, the end-user is only intended to calculate the
response dispersion.}

\subsection{Grading the LLM responses to the IRC-Wiki Trivia
questions}\label{grading-the-llm-responses-to-the-irc-wiki-trivia-questions}

In total there were 13888 responses to be graded across the different questions and models.\footnote{One might expect 10 models * 1397 to mean there would be 13970 responses to grade, however occassionally requests would be dropped or otherwise recieve some internal server error as a response. Since this was spread roughly evenly across models and question categories it will not bias the dataset.} In order to evaluate
possible automated grading methodologies, the author sampled 1000 responses from
the 13888 responses and hand-graded whether, the author being
informed by the trivia question and the answer key's answer, if the sampled
responses contained the answer key's answer within it. The goal was to
have the hand-graded sample be a gold standard in capturing subtleties in responses that mere
string comparisons could easily miss, such as when answering the question "Which two countries share Victoria Falls?" the answer "zambia and zimbabwe" provided by an LLM being actually the same as the
answer key's "zimbabwe and zambia" or that in answer the question "In which film did Bill Murray drive an ectomobile?" the LLM's response "the answer is ghostbusters 1984" is marked correct against the answer key's "ghostbusters". Thus each of the 1000
sampled responses was hand-checked and graded that the LLM in that case
had gotten the answer right, wrong, or if it was ambigous to the author
whether the LLM's response should be considered right or wrong then it
was held out. 30 responses out of the 1000 sampled fell into this last
category, and so 970 hand-checked grades were used in evaluating
automated grading methodologies.

The first methodology evaluated for autograding the LLM responses was to
remove punctuation from the answer key's answer and LLM response,
lowercase all letters, and check if the post-processed answer key's
answer was a substring of the post-processed LLM response. Based on the
970 hand-graded examples, this assigned the correct grade to the LLM's
response 90.5\% of the time.

The second grading methodology was to grade each LLM response by sending
the following prompt to GPT4 for each question, answer key answer, and
LLM response:

\begin{quote}
Here is a test for evaluating LLMs. I want to see how well you follow my instructions when constructing your response.

I want you to respond with a SINGLE WORD, either "Yes" or "No", and ONLY THAT. Do not add any other context, other words, notes, explanations, justifications, or objections to my phrasing of this question. 
Now, I am going to show you a trivia question in the category "\{category\}". I am going to show you the question, and then the answer key's answer, and then the answer from an LLM respondent. I want you to respond "Yes" if the *respondent's* response contains the correct answer. Otherwise, answer "No". 
Here is the question: "\{question\}"  
Here is the the answer key's answer: "\{answer\_key\_answer\}"
Here is the responent's response: "\{response\}"
Now please respond with just Yes/No.
\end{quote}

GPT4 successfully followed the prompt's instruction to always answer
with just ``Yes'' or ``No''. GPT4's grading agreed with the author on
98.4\% the hand-graded sample, so GPT4's grades are used for the rest of
this paper.

    \subsection{Response Dispersion Use-Case Analysis
Defined}\label{response-dispersion-use-case-analysis-defined}

    For each possible tolerance parameter, every unique model-pair and topic
category combination goes through the following process:

\begin{enumerate}
\def\labelenumi{\arabic{enumi}.}
\item
  The response dispersion is calculated for each of the two models on
  that category. The model with the lower response dispersion is called
  ``the chosen model'' in this process.
\item
  If, within this category, the chosen model has a ground truth accuracy
  that is greater than the other model's ground truth accuracy minus the
  tolerance level, then this is scored as a success for the process.
\end{enumerate}

The success \% of outcomes is calculated per category, since models are
only intended to be compared within a given domain of discourse, same as
hand-labeled qa datasets would only be defined over a particular domain
of discouse of interest.

A baseline for each category is also defined for comparison. It is
calculated using a 100 iteration monte carlo simulation where in one
iteration every unique pair of models is compared:

\begin{enumerate}
\def\labelenumi{\arabic{enumi}.}
\item
  ``The chosen model'' in this case is chosen at random.
\item
  If, within this category, the chosen model has a ground truth accuracy
  that is greater than the other model's ground truth accuracy minus the
  tolerance level, then this is scored as a success.
\end{enumerate}

The success \% of outcomes for the baseline is calculated per category, averaged over the
100 iterations.

\subsection{LLMs studied}\label{llms-studied}

This research studied 10 candidate models in particular, all of whom were queried through the Openrouter.ai API. The following table lists the Openrouter.ai model identifier (which is a cannonical identifier for the model and model-version used) and the short name for the model used later in the results section of this paper. 
\begin{center}
\begin{tabular}{| l | l |}
\hline
\textbf{Openrouter.ai model identifier} & \textbf{Short name} \\ \hline
01-ai/yi-34b-chat & Yi-34b \\ \hline
anthropic/claude-3-opus & Claude3-Opus \\ \hline
codellama/codellama-70b-instruct & Codellama-70b \\ \hline
google/gemini-pro & Gemini-Pro \\ \hline
gpt-3.5-turbo-1106 & GPT-3.5 \\ \hline
gpt-4-1106-preview & GPT-4 \\ \hline
meta-llama/llama-2-70b-chat & Llama2-70b \\ \hline
meta-llama/llama-3-70b-instruct:nitro & Llama3-70b \\ \hline
meta-llama/llama-3-8b-instruct:nitro & Llama3-8b \\ \hline
mistralai/mistral-7b-instruct & Mistral-7bv0.1 \\ \hline
mistralai/mistral-7b-instruct:nitro & Mistral-7bv0.2 \\ \hline
mistralai/mixtral-8x22b & Mixtral-8x22b \\ \hline
mistralai/mixtral-8x7b-instruct:nitro & Mixtral-8x7b \\ \hline
\end{tabular}
\end{center}
    
    \section{Results}\label{results}

    \subsection{Success \% averaged over all
categories}\label{success-averaged-over-all-categories}

\begin{tabular}{lccc}
\hline
Response Dispersion by Embedding & 0\% tolerance & 5\% tolerance & 10\% tolerance \\ \hline
Reference Sentence Similarities (RSS)        & 71.5\%  & 80.4\% & 87.2\% \\
OpenAI \texttt{text-embedding-3-large}        & 73.5\%  & 81.7\%  & 88.5\% \\ \hline
random choice baseline    & 50.9\%  & 59.5\%  & 67.5\% \\
\hline
\end{tabular}

\begin{figure}
    \centering
    \includegraphics[width=1.0\linewidth]{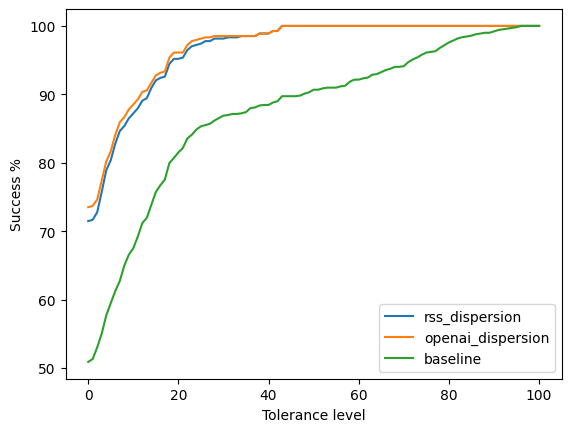}
    \caption{Success \% of choosing the best or good-enough model at different tolerance levels, averaged over all models and topic categories}
    \label{fig:enter-label}
\end{figure}

The success \% averaged over all categories is graphed against all tolerance levels in \autoref{fig:enter-label}.

\clearpage

\subsection{Models ranked in each category by QA Benchmark, and Response Dispersion}

The following tables give details, broken down by each topic category found in \texttt{IRC-WikiTriviaQA}, of the accuracy of each model on the QA benchmark, of the orderings of the models given by rank dispersion calculations for both the OpenAI \texttt{text-embedding-3-large} embeddings and the RSS embeddings, and the correlations between the rankings of the models by all three metrics.

\begin{longtable}{| m{5cm} | m{12cm} |}
\hline
\multicolumn{2}{|l|}{\textbf{Category:Animals}} \\
\multicolumn{2}{|l|}{Spearman rank correlations: QA-OpenAI -0.59; QA-RSS -0.59; RSS-OpenAI 0.98} \\ \hline
QA Benchmark & 1. Claude3-Opus (91.4\%), 2. GPT-4 (88.6\%), 3. GPT-3.5 (85.7\%), 4. Gemini-Pro (82.1\%), 5. Yi-34b (79.3\%), 6. Mixtral-8x7b (77.9\%), 7. Llama2-70b (74.3\%), 8. Mistral-7bv0.1 (68.6\%), 9. Mistral-7bv0.2 (67.9\%), 10. Codellama-70b (11.4\%) \\ \hline
OpenAI's \texttt{text-embedding-3-large} Response Dispersion & 1. Claude3-Opus (5), 2. GPT-4 (6), 3. Mistral-7bv0.2 (15), 4. GPT-3.5 (16), 5. Mistral-7bv0.1 (25), 6. Gemini-Pro (29), 7. Llama2-70b (42), 8. Yi-34b (60), 9. Mixtral-8x7b (68), 10. Codellama-70b (81) \\ \hline
RSS Response Dispersion & 1. GPT-4 (3), 2. Claude3-Opus (4), 3. Mistral-7bv0.2 (7), 4. GPT-3.5 (8), 5. Gemini-Pro (14), 6. Mistral-7bv0.1 (14), 7. Llama2-70b (23), 8. Yi-34b (47), 9. Mixtral-8x7b (47), 10. Codellama-70b (79) \\ \hline
\multicolumn{2}{|l|}{\textbf{Category:Computers}} \\
\multicolumn{2}{|l|}{Spearman rank correlations: QA-OpenAI -0.64; QA-RSS -0.64; RSS-OpenAI 0.98} \\ \hline
QA Benchmark & 1. Claude3-Opus (86.2\%), 2. GPT-4 (82.8\%), 3. GPT-3.5 (82.8\%), 4. Yi-34b (74.1\%), 5. Llama2-70b (69.0\%), 6. Mixtral-8x7b (65.5\%), 7. Gemini-Pro (65.5\%), 8. Mistral-7bv0.1 (58.6\%), 9. Mistral-7bv0.2 (55.2\%), 10. Codellama-70b (3.5\%) \\ \hline
OpenAI's \texttt{text-embedding-3-large} Response Dispersion & 1. Claude3-Opus (2), 2. GPT-4 (3), 3. Mixtral-8x7b (3), 4. GPT-3.5 (9), 5. Llama2-70b (12), 6. Mistral-7bv0.2 (12), 7. Mistral-7bv0.1 (18), 8. Gemini-Pro (23), 9. Yi-34b (58), 10. Codellama-70b (85) \\ \hline
RSS Response Dispersion & 1. Claude3-Opus (1), 2. GPT-4 (1), 3. Mixtral-8x7b (1), 4. GPT-3.5 (5), 5. Llama2-70b (5), 6. Mistral-7bv0.2 (5), 7. Mistral-7bv0.1 (9), 8. Gemini-Pro (12), 9. Yi-34b (47), 10. Codellama-70b (84) \\ \hline
\multicolumn{2}{|l|}{\textbf{Category:Food}} \\
\multicolumn{2}{|l|}{Spearman rank correlations: QA-OpenAI -0.54; QA-RSS -0.54; RSS-OpenAI 0.98} \\ \hline
QA Benchmark & 1. Claude3-Opus (91.2\%), 2. GPT-4 (87.7\%), 3. GPT-3.5 (82.9\%), 4. Mixtral-8x7b (75.4\%), 5. Gemini-Pro (74.6\%), 6. Yi-34b (74.2\%), 7. Llama2-70b (71.5\%), 8. Mistral-7bv0.1 (68.4\%), 9. Mistral-7bv0.2 (65.8\%), 10. Codellama-70b (1.8\%) \\ \hline
OpenAI's \texttt{text-embedding-3-large} Response Dispersion & 1. Claude3-Opus (4), 2. Mistral-7bv0.2 (5), 3. GPT-4 (6), 4. Mixtral-8x7b (7), 5. GPT-3.5 (9), 6. Mistral-7bv0.1 (14), 7. Llama2-70b (19), 8. Gemini-Pro (27), 9. Yi-34b (64), 10. Codellama-70b (82) \\ \hline
RSS Response Dispersion & 1. Mistral-7bv0.2 (1), 2. Claude3-Opus (2), 3. GPT-4 (2), 4. Mixtral-8x7b (3), 5. GPT-3.5 (5), 6. Mistral-7bv0.1 (7), 7. Llama2-70b (10), 8. Gemini-Pro (12), 9. Yi-34b (52), 10. Codellama-70b (81) \\ \hline
\multicolumn{2}{|l|}{\textbf{Category:Football}} \\
\multicolumn{2}{|l|}{Spearman rank correlations: QA-OpenAI -0.79; QA-RSS -0.79; RSS-OpenAI 0.99} \\ \hline
QA Benchmark & 1. GPT-4 (100.0\%), 2. Claude3-Opus (98.8\%), 3. GPT-3.5 (98.2\%), 4. Gemini-Pro (95.3\%), 5. Llama2-70b (91.1\%), 6. Mixtral-8x7b (86.4\%), 7. Yi-34b (82.0\%), 8. Mistral-7bv0.2 (79.3\%), 9. Mistral-7bv0.1 (77.5\%), 10. Codellama-70b (4.7\%) \\ \hline
OpenAI's \texttt{text-embedding-3-large} Response Dispersion & 1. Claude3-Opus (4), 2. GPT-4 (5), 3. GPT-3.5 (6), 4. Gemini-Pro (19), 5. Mistral-7bv0.2 (23), 6. Mistral-7bv0.1 (27), 7. Llama2-70b (39), 8. Yi-34b (59), 9. Mixtral-8x7b (79), 10. Codellama-70b (83) \\ \hline
RSS Response Dispersion & 1. GPT-4 (2), 2. Claude3-Opus (3), 3. GPT-3.5 (4), 4. Gemini-Pro (8), 5. Mistral-7bv0.2 (13), 6. Mistral-7bv0.1 (18), 7. Llama2-70b (22), 8. Yi-34b (48), 9. Mixtral-8x7b (64), 10. Codellama-70b (83) \\ \hline
\multicolumn{2}{|l|}{\textbf{Category:Geography}} \\
\multicolumn{2}{|l|}{Spearman rank correlations: QA-OpenAI -0.49; QA-RSS -0.49; RSS-OpenAI 0.99} \\ \hline
QA Benchmark & 1. GPT-3.5 (93.5\%), 2. Claude3-Opus (92.2\%), 3. GPT-4 (90.9\%), 4. Gemini-Pro (87.0\%), 5. Yi-34b (86.8\%), 6. Llama2-70b (84.4\%), 7. Mixtral-8x7b (83.1\%), 8. Mistral-7bv0.2 (80.5\%), 9. Mistral-7bv0.1 (80.5\%), 10. Codellama-70b (13.0\%) \\ \hline
OpenAI's \texttt{text-embedding-3-large} Response Dispersion & 1. Claude3-Opus (4), 2. GPT-4 (5), 3. Mistral-7bv0.2 (15), 4. Llama2-70b (17), 5. GPT-3.5 (18), 6. Mistral-7bv0.1 (19), 7. Gemini-Pro (20), 8. Mixtral-8x7b (26), 9. Yi-34b (67), 10. Codellama-70b (85) \\ \hline
RSS Response Dispersion & 1. Claude3-Opus (3), 2. GPT-4 (3), 3. Llama2-70b (7), 4. Mistral-7bv0.2 (7), 5. GPT-3.5 (8), 6. Mistral-7bv0.1 (8), 7. Gemini-Pro (10), 8. Mixtral-8x7b (18), 9. Yi-34b (54), 10. Codellama-70b (84) \\ \hline
\multicolumn{2}{|l|}{\textbf{Category:History}} \\
\multicolumn{2}{|l|}{Spearman rank correlations: QA-OpenAI -0.69; QA-RSS -0.69; RSS-OpenAI 0.99} \\ \hline
QA Benchmark & 1. Claude3-Opus (97.6\%), 2. GPT-4 (94.0\%), 3. GPT-3.5 (91.7\%), 4. Gemini-Pro (86.9\%), 5. Yi-34b (83.1\%), 6. Mixtral-8x7b (81.0\%), 7. Mistral-7bv0.2 (81.0\%), 8. Mistral-7bv0.1 (79.8\%), 9. Llama2-70b (79.8\%), 10. Codellama-70b (2.4\%) \\ \hline
OpenAI's \texttt{text-embedding-3-large} Response Dispersion & 1. GPT-4 (3), 2. Mixtral-8x7b (4), 3. Claude3-Opus (5), 4. GPT-3.5 (15), 5. Mistral-7bv0.2 (16), 6. Gemini-Pro (19), 7. Mistral-7bv0.1 (24), 8. Llama2-70b (46), 9. Yi-34b (72), 10. Codellama-70b (84) \\ \hline
RSS Response Dispersion & 1. GPT-4 (1), 2. Mixtral-8x7b (1), 3. Claude3-Opus (2), 4. GPT-3.5 (7), 5. Mistral-7bv0.2 (7), 6. Gemini-Pro (8), 7. Mistral-7bv0.1 (14), 8. Llama2-70b (25), 9. Yi-34b (63), 10. Codellama-70b (84) \\ \hline
\multicolumn{2}{|l|}{\textbf{Category:Movies}} \\
\multicolumn{2}{|l|}{Spearman rank correlations: QA-OpenAI -0.55; QA-RSS -0.55; RSS-OpenAI 0.96} \\ \hline
QA Benchmark & 1. Claude3-Opus (90.6\%), 2. GPT-4 (88.3\%), 3. GPT-3.5 (80.3\%), 4. Gemini-Pro (70.4\%), 5. Yi-34b (69.3\%), 6. Mixtral-8x7b (66.4\%), 7. Llama2-70b (66.4\%), 8. Mistral-7bv0.1 (53.4\%), 9. Mistral-7bv0.2 (52.0\%), 10. Codellama-70b (17.5\%) \\ \hline
OpenAI's \texttt{text-embedding-3-large} Response Dispersion & 1. Claude3-Opus (5), 2. Mixtral-8x7b (5), 3. GPT-4 (6), 4. GPT-3.5 (13), 5. Mistral-7bv0.2 (20), 6. Mistral-7bv0.1 (25), 7. Gemini-Pro (27), 8. Llama2-70b (52), 9. Yi-34b (63), 10. Codellama-70b (84) \\ \hline
RSS Response Dispersion & 1. GPT-4 (3), 2. Mixtral-8x7b (3), 3. Claude3-Opus (4), 4. GPT-3.5 (7), 5. Mistral-7bv0.2 (10), 6. Gemini-Pro (12), 7. Mistral-7bv0.1 (15), 8. Llama2-70b (31), 9. Yi-34b (52), 10. Codellama-70b (81) \\ \hline
\multicolumn{2}{|l|}{\textbf{Category:Music}} \\
\multicolumn{2}{|l|}{Spearman rank correlations: QA-OpenAI -0.1; QA-RSS -0.1; RSS-OpenAI 0.99} \\ \hline
QA Benchmark & 1. GPT-4 (87.4\%), 2. Claude3-Opus (85.6\%), 3. GPT-3.5 (82.4\%), 4. Yi-34b (67.2\%), 5. Gemini-Pro (66.0\%), 6. Llama2-70b (62.5\%), 7. Mixtral-8x7b (59.8\%), 8. Mistral-7bv0.2 (44.7\%), 9. Mistral-7bv0.1 (44.7\%), 10. Codellama-70b (4.1\%) \\ \hline
OpenAI's \texttt{text-embedding-3-large} Response Dispersion & 1. Mixtral-8x7b (2), 2. GPT-4 (5), 3. Mistral-7bv0.2 (5), 4. Mistral-7bv0.1 (7), 5. GPT-3.5 (13), 6. Claude3-Opus (15), 7. Gemini-Pro (26), 8. Llama2-70b (36), 9. Yi-34b (63), 10. Codellama-70b (85) \\ \hline
RSS Response Dispersion & 1. Mixtral-8x7b (1), 2. Mistral-7bv0.2 (2), 3. GPT-4 (3), 4. Mistral-7bv0.1 (3), 5. GPT-3.5 (7), 6. Claude3-Opus (8), 7. Gemini-Pro (13), 8. Llama2-70b (21), 9. Yi-34b (52), 10. Codellama-70b (79) \\ \hline
\multicolumn{2}{|l|}{\textbf{Category:Science}} \\
\multicolumn{2}{|l|}{Spearman rank correlations: QA-OpenAI -0.71; QA-RSS -0.71; RSS-OpenAI 1.0} \\ \hline
QA Benchmark & 1. Claude3-Opus (94.6\%), 2. GPT-4 (92.5\%), 3. GPT-3.5 (89.3\%), 4. Gemini-Pro (87.5\%), 5. Mixtral-8x7b (83.9\%), 6. Yi-34b (80.0\%), 7. Llama2-70b (76.8\%), 8. Mistral-7bv0.2 (73.2\%), 9. Mistral-7bv0.1 (73.2\%), 10. Codellama-70b (1.8\%) \\ \hline
OpenAI's \texttt{text-embedding-3-large} Response Dispersion & 1. Claude3-Opus (3), 2. Mixtral-8x7b (3), 3. GPT-4 (7), 4. GPT-3.5 (12), 5. Mistral-7bv0.2 (18), 6. Gemini-Pro (24), 7. Mistral-7bv0.1 (24), 8. Llama2-70b (31), 9. Yi-34b (61), 10. Codellama-70b (85) \\ \hline
RSS Response Dispersion & 1. Claude3-Opus (1), 2. Mixtral-8x7b (2), 3. GPT-4 (4), 4. GPT-3.5 (7), 5. Mistral-7bv0.2 (8), 6. Gemini-Pro (12), 7. Mistral-7bv0.1 (12), 8. Llama2-70b (17), 9. Yi-34b (50), 10. Codellama-70b (85) \\ \hline
\multicolumn{2}{|l|}{\textbf{Category:Sport}} \\
\multicolumn{2}{|l|}{Spearman rank correlations: QA-OpenAI -0.81; QA-RSS -0.81; RSS-OpenAI 0.99} \\ \hline
QA Benchmark & 1. GPT-4 (92.9\%), 2. Mixtral-8x7b (85.7\%), 3. Claude3-Opus (85.7\%), 4. GPT-3.5 (78.6\%), 5. Gemini-Pro (78.6\%), 6. Yi-34b (78.6\%), 7. Llama2-70b (57.1\%), 8. Mistral-7bv0.2 (35.7\%), 9. Mistral-7bv0.1 (35.7\%), 10. Codellama-70b (21.4\%) \\ \hline
OpenAI's \texttt{text-embedding-3-large} Response Dispersion & 1. Claude3-Opus (6), 2. GPT-4 (7), 3. Mixtral-8x7b (12), 4. GPT-3.5 (14), 5. Llama2-70b (20), 6. Gemini-Pro (24), 7. Mistral-7bv0.2 (24), 8. Mistral-7bv0.1 (42), 9. Yi-34b (63), 10. Codellama-70b (83) \\ \hline
RSS Response Dispersion & 1. Claude3-Opus (3), 2. GPT-4 (4), 3. GPT-3.5 (6), 4. Mixtral-8x7b (6), 5. Gemini-Pro (11), 6. Llama2-70b (11), 7. Mistral-7bv0.2 (13), 8. Mistral-7bv0.1 (27), 9. Yi-34b (49), 10. Codellama-70b (82) \\ \hline
\multicolumn{2}{|l|}{\textbf{Category:TV}} \\
\multicolumn{2}{|l|}{Spearman rank correlations: QA-OpenAI -0.57; QA-RSS -0.57; RSS-OpenAI 1.0} \\ \hline
QA Benchmark & 1. GPT-4 (94.1\%), 2. Claude3-Opus (94.1\%), 3. Mixtral-8x7b (82.3\%), 4. GPT-3.5 (82.3\%), 5. Llama2-70b (70.6\%), 6. Yi-34b (70.6\%), 7. Mistral-7bv0.2 (64.7\%), 8. Mistral-7bv0.1 (64.7\%), 9. Gemini-Pro (58.8\%), 10. Codellama-70b (23.5\%) \\ \hline
OpenAI's \texttt{text-embedding-3-large} Response Dispersion & 1. Claude3-Opus (5), 2. GPT-4 (11), 3. GPT-3.5 (17), 4. Mistral-7bv0.2 (26), 5. Gemini-Pro (28), 6. Mistral-7bv0.1 (31), 7. Llama2-70b (54), 8. Yi-34b (63), 9. Mixtral-8x7b (82), 10. Codellama-70b (83) \\ \hline
RSS Response Dispersion & 1. Claude3-Opus (2), 2. GPT-4 (5), 3. GPT-3.5 (8), 4. Mistral-7bv0.2 (14), 5. Gemini-Pro (15), 6. Mistral-7bv0.1 (18), 7. Llama2-70b (27), 8. Yi-34b (50), 9. Mixtral-8x7b (69), 10. Codellama-70b (83) \\ \hline
\multicolumn{2}{|l|}{\textbf{Category:TV-Cartoons}} \\
\multicolumn{2}{|l|}{Spearman rank correlations: QA-OpenAI -0.64; QA-RSS -0.64; RSS-OpenAI 0.98} \\ \hline
QA Benchmark & 1. GPT-4 (87.0\%), 2. Claude3-Opus (84.1\%), 3. GPT-3.5 (81.2\%), 4. Mixtral-8x7b (72.5\%), 5. Gemini-Pro (72.5\%), 6. Llama2-70b (68.1\%), 7. Yi-34b (68.1\%), 8. Mistral-7bv0.1 (55.1\%), 9. Mistral-7bv0.2 (53.6\%), 10. Codellama-70b (7.2\%) \\ \hline
OpenAI's \texttt{text-embedding-3-large} Response Dispersion & 1. Claude3-Opus (6), 2. GPT-3.5 (6), 3. GPT-4 (7), 4. Mistral-7bv0.2 (10), 5. Llama2-70b (17), 6. Gemini-Pro (18), 7. Mistral-7bv0.1 (22), 8. Yi-34b (66), 9. Mixtral-8x7b (81), 10. Codellama-70b (84) \\ \hline
RSS Response Dispersion & 1. GPT-3.5 (2), 2. Claude3-Opus (4), 3. GPT-4 (4), 4. Mistral-7bv0.2 (4), 5. Llama2-70b (5), 6. Gemini-Pro (8), 7. Mistral-7bv0.1 (9), 8. Yi-34b (56), 9. Mixtral-8x7b (69), 10. Codellama-70b (82) \\ \hline

\end{longtable}

\section{Discussion and Future Work}\label{discussion-and-future}

\subsection{Shortcomings}

This document repurposed a trivia question dataset previously unused for QA benchmarking. While there are many already-existing QA benchmark datasets (see, for example, the long but non-exhaustive list in \citet{wang_modern_2022} and \citet[section 4]{chang_survey_2024}), few of them attach explicit topic categories to their respective questions. One of the rare examples to do so subset of the Quasar dataset \citep{dhingra_quasar_2017} called \texttt{Quasar-T-dev}, which contains 132 questions over 9 topic categories (math \& science 21, arts 10, language 15, food 14, movies \& music 36, sports 3, general 24, history \& religion 33, people \& places 58; note the sum is greater than 132 questions because some questions belong in multiple categories). There are, however, some QA datasets intended for specific topic domains and so they can provide at least another data point, such as \texttt{MultiMedQA}, "a benchmark combining six existing medical question answering datasets spanning professional medicine, research and consumer queries and a new dataset of medical questions searched online, \texttt{HealthSearchQA}", from \citep{singhal_large_2023}. It is also study to dropped categories from the original IRC-Wiki Trivia dataset for more datapoints. The author originally dropped them to keep the studied categories roughly comparable in terms of representing general-knowledge trivia categories so that the averages calculated in subsection \ref{success-averaged-over-all-categories} wouldn't be weighted too heavily by niche topics, however that doesn't mean they wouldn't be useful datapoints to have in a larger study.

\subsection{Future Directions}

Since response dispersion is an automated way to gauge topic knowledge, the author feels it would be useful as another automated metric to track when finetuning LLMs for a specific task. Thus response dispersion can be used to track an LLM learning about a task. Furthermore, when incrementally learning from new data, LLMs are at risk of "catastrophic forgetting" of prior probability distributions as its weights change \citep{van_de_ven_three_2022}. The author believes and would like to see validation in the future that measuring response dispersion can be metric that guards against catastrophic forgetting of tracked topics, since an increase in response dispersion for an already-known topic should indicate a loss of certainty in the LLM's response generations about that topic.

This paper also introduces reference sentence similarity embeddings and shows that the response dispersion defined by them performs nearly as well for comparing hypothetical QA benchmarks as do the embeddings produced by OpenAI's \texttt{text-embedding-3-large} text embeddings, all while being fast to compute locally. RSS embeddings seem to make a lot of sense in any context where you are comparing one sentence against other already known relevant sentences. This strongly suggests, theoretically at least, that rss embeddings should perform well for Retrieval Augmented Generation (RAG), where the task is to find documents relevant to a given user query in order to generate a relevant answer from the relevant documents.

\bibliography{references} 

\end{document}